%% file: 0_main.tex
\documentclass[letterpaper, 10 pt, conference]{ieeeconf}  % Comment this line out if you need a4paper

\IEEEoverridecommandlockouts                              % This command is only needed if 
\usepackage{amsmath} % assumes amsmath package installed
\usepackage{amssymb}  % assumes amsmath package installed
\usepackage{graphicx}
\usepackage{color}
\usepackage{subfig}

\usepackage{algorithm}
\usepackage{algorithmicx}
\usepackage{algpseudocode}

\usepackage{url}
\let\OldUrlFont\UrlFont 
\renewcommand{\UrlFont}{\small\OldUrlFont}

\usepackage[detect-none]{siunitx}
\newcommand{\argmax}{\arg\!\max}

\newcommand{\vs}[0]{\vec{s}}
\newcommand{\vo}[0]{\vec{o}}
\newcommand{\vr}[0]{\vec{r}}
\newcommand{\va}[0]{\vec{a}}

\title{\LARGE \bf
Learning to Represent Haptic Feedback for Partially-Observable Tasks
}

\author{Jaeyong Sung$^{1,2}$, J. Kenneth Salisbury$^{1}$ and Ashutosh Saxena$^{3}$% <-this % stops a space
\thanks{$^{1}$Department of Computer Science, Stanford University.
        $^{2}$Department of Computer Science, Cornell University.
        $^{3}$Brain Of Things, Inc.
        {\tt\small \{jysung,jks,asaxena\}@cs.stanford.edu}}%        
}

\begin{document}

\maketitle
\thispagestyle{empty}
\pagestyle{empty}

\maketitle

%%%%%%%%%%%%%%%%%%%%%%%%%%%%%%%%%%%%%%%%%%%%%%%%%%%%%%%%%%%%%%%%%%%%%%%%%%%%%%%%
\begin{abstract}
The sense of touch, 
being the earliest sensory system to develop in a human body \cite{montagu1971touching},
plays a critical part of our daily interaction with the environment.
In order to successfully complete a task,
many manipulation interactions require incorporating haptic feedback.
However, manually designing a feedback mechanism
can be extremely challenging.
In this work, we consider manipulation tasks that need to
incorporate tactile sensor feedback in order to modify a provided nominal plan.
To incorporate partial observation, we present a new framework that 
models the task as a partially observable Markov decision process (POMDP) 
and learns an appropriate representation of haptic feedback which 
can serve as the state for a POMDP model.
The model, that is parametrized by deep recurrent neural networks,
utilizes variational Bayes methods to optimize the approximate posterior.
Finally, we build on deep Q-learning to be able to select the optimal
action in each state without access to a simulator.
We test our model on a PR2 robot 
for multiple tasks of turning a knob until it clicks.
\end{abstract}

%%%%%%%%%%%%%%%%%%%%%%%%%%%%%%%%%%%%%%%%%%%%%%%%%%%%%%%%%%%%%%%%%%%%%%%%%%%%%%%%
\input{space_saver}

\input{1_introduction}

\input{2_relatedwork}

\input{3_approach}

\input{4_results}

\input{5_conclusion}

%%%%%%%%%%%%%%%%%%%
%\section*{APPENDIX}
\input{6_appendix}

%%%%%%%%%%%%%%%%%%%%%%%
%\section*{Acknowledgments}
%%%%%%%%%%%%%%%%%%%%%%%%%%%%%
%\vspace*{-0.03in}
%\section*{ACKNOWLEDGMENT}
%\vspace*{-0.03in}

\vspace*{.055in}
\noindent
\textbf{Acknowledgment.}
We thank Ian Lenz for useful discussions. 
This work was supported 
by Microsoft Faculty Fellowship and NSF Career Award to Saxena.

%%%%%%%%%%%%%%%%%%%%%%%
{
%\vspace*{-.05in}
\baselineskip 0cm
\bibliographystyle{IEEEtran}
\bibliography{references}
}

\end{document}

%% file: space_saver.tex
% Some space-saving macros
% \parskip=5pt
  \abovedisplayskip 3.0pt plus2pt minus2pt%
 \belowdisplayskip \abovedisplayskip
\renewcommand{\baselinestretch}{0.97}

\newenvironment{packed_enum}{
\begin{enumerate}
  \setlength{\itemsep}{0pt}
  \setlength{\parskip}{0pt}
  \setlength{\parsep}{0pt}
}
{\end{enumerate}}

\newenvironment{packed_item}{
\begin{itemize}
  \setlength{\itemsep}{0pt}
  \setlength{\parskip}{0pt}
  \setlength{\parsep}{0pt}
}{\end{itemize}}

\newlength\savedwidth
\newcommand\whline[1]{\noalign{\global\savedwidth\arrayrulewidth
                               \global\arrayrulewidth #1} %
                      \hline
                      \noalign{\global\arrayrulewidth\savedwidth}}

\newlength{\sectionReduceTop}
\newlength{\sectionReduceBot}
\newlength{\subsectionReduceTop}
\newlength{\subsectionReduceBot}
\newlength{\abstractReduceTop}
\newlength{\abstractReduceBot}
\newlength{\captionReduceTop}
\newlength{\captionReduceBot}
%\newlength{\nameReduceTop}
\newlength{\subsubsectionReduceTop}
\newlength{\subsubsectionReduceBot}
\newlength{\headerReduceTop}
% Negative space for figures set at the bottom of a block of figs
\newlength{\figureReduceBot}

\newlength{\horSkip}
\newlength{\verSkip}

\newlength{\equationReduceTop}

\newlength{\figureHeight}
\setlength{\figureHeight}{1.7in}

%\newlength{\figureFraction}
\setlength{\horSkip}{-.09in}
\setlength{\verSkip}{-.1in}
%\setlength{\figureFraction}{.195}

% figureReduceBot is for figures which are set above text, since latex
% likes putting a lot of space under those
\setlength{\figureReduceBot}{0in}
\setlength{\headerReduceTop}{0in}
\setlength{\subsectionReduceTop}{0in}
\setlength{\subsectionReduceBot}{-0.03in}
\setlength{\sectionReduceTop}{-0.0in}
\setlength{\sectionReduceBot}{-0.01in}

\setlength{\subsubsectionReduceTop}{0in}
\setlength{\subsubsectionReduceBot}{0in}
\setlength{\abstractReduceTop}{0in}
\setlength{\abstractReduceBot}{0in}

\setlength{\equationReduceTop}{0in}

\setlength{\captionReduceTop}{0in}
\setlength{\captionReduceBot}{0in}

%% file: 1_introduction.tex
\section{Introduction}
\label{sec:intro}

\begin{figure}[t]
  \begin{center}
    \includegraphics[width=\columnwidth,trim={0cm 0cm 0cm 0cm}, clip]{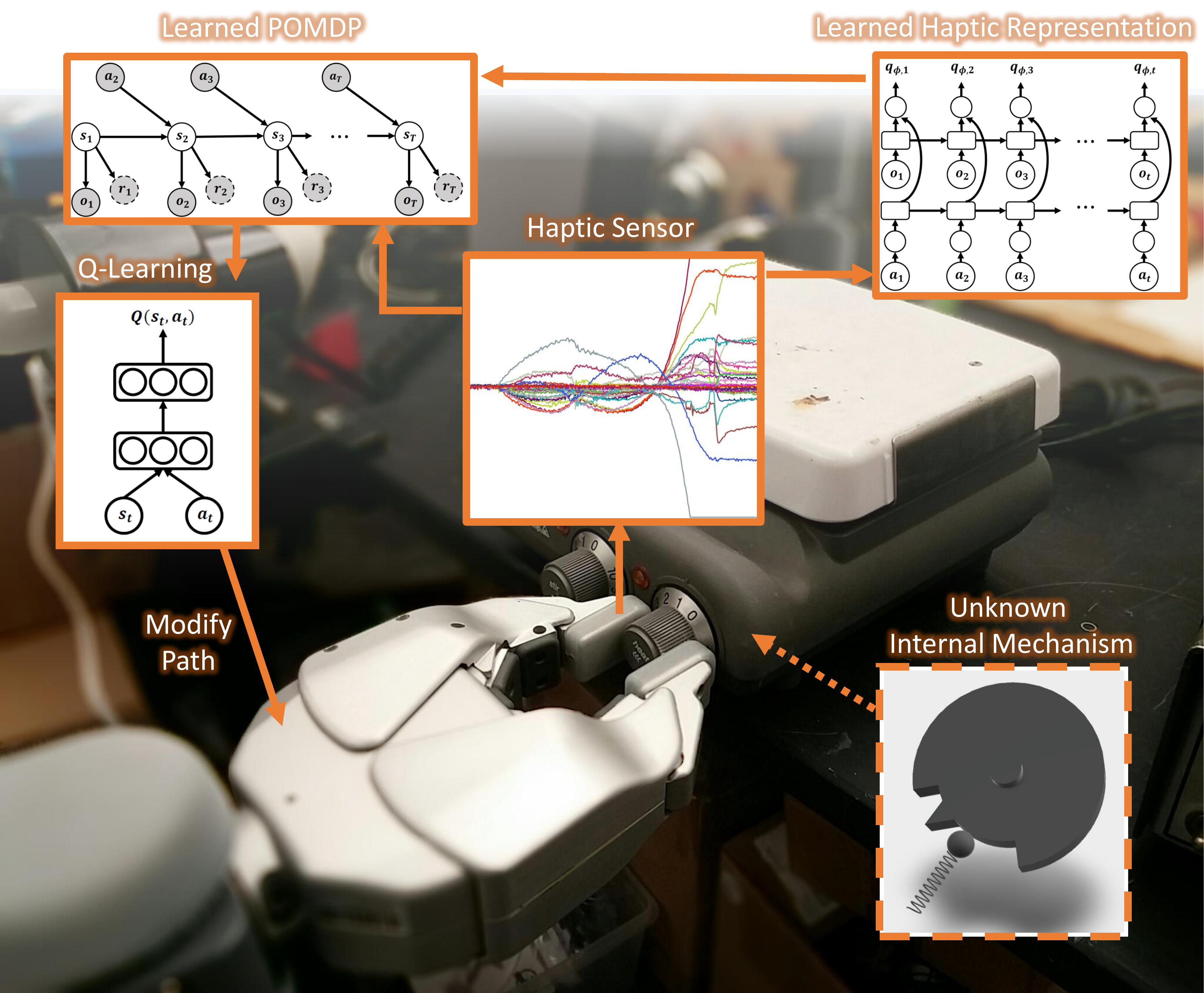}
  \vskip -0.15in
  \end{center}
  \caption{
    %\textbf{PR2 Manipulating Clicking Knob} with haptic (tactile) feedback at its fingertips.
    %Learning to represent haptic feedback to rotate the knob until it clicks.
    \textbf{Haptic feedback} from a tactile sensor being used to modify a nominal plan of manipulation.
    Our framework learns an appropriate representation (embedding space) which in turn is used to learn 
    to find optimal control. 
    }
  \label{fig:main_fig}
\end{figure}

Many tasks in human environments that we do without much effort
require more than just visual observation. Very often they require
incorporating the sense of touch to complete the task.
For example, consider the task of turning a knob that needs to be rotated
until it clicks, like the one in Figure~\ref{fig:main_fig}.
The robot could observe the consequence of its action if any visible changes occur,
but such clicks can often only be directly observed through the fingers.
Many of the objects that surround us are explicitly designed with feedback --- one of the key interaction design principles --- otherwise ``one is always wondering whether anything has happened'' \cite{norman1988design}.

Recently, there has been a lot of progress in making
robots understand and act based on images \cite{levine2015end,watter2015embed,mnih2013playing} 
and point-clouds \cite{sung_robobarista_2015}.
A robot can definitely gain a lot of information from visual sensors,
including a nominal trajectory plan for a task \cite{sung_robobarista_2015}.
However, when the robot is manipulating a small object or 
once the robot starts interacting with small parts of appliances,
self-occlusion by its own arms and its end-effectors
limits the use of the visual information.

However, building an algorithm that can examine haptic properties 
and incorporate such information to influence a motion is
very challenging for multiple reasons.
First, haptic feedback  is a dynamic response that is dependent
on the action the robot has taken on the object as well as internal
states and properties of the object.
Second, every haptic sensor produces a
vastly different raw sensor signal.

Moreover, compared to the rich information that can be extracted about a current state of the task
from few images 
(e.g. position and velocity information of an end-effector and an object \cite{mnih2013playing,levine2015end}), 
a short window of haptic sensor signal is merely a partial consequence of 
the interaction and of the changes in an unobservable internal mechanism.
It also suffers from perceptual aliasing --- \textit{i.e.} many segments of a haptic signal
at different points of interaction can produce a very similar signal.
These challenges make it difficult to design an algorithm that
can incorporate information from haptic modalities (in our case, tactile sensors).

In this work, we introduce a framework that can learn to represent haptic
feedback for tasks requiring incorporation of a haptic signal.
Since a haptic signal only provides a partial observation,
we model the task using a partially observable Markov decision process (POMDP).
However, since we do not know of definition of states for a POMDP,
we first learn an appropriate representation from a haptic signal to be used as continuous states for a POMDP.
To overcome the intractability in computing the posterior,
we employ a variational Bayesian method, with a deep recurrent neural network,
that maximizes lower bound of likelihood of the training data.

%\todo{QLearning}
Using a learned representation of the interaction with feedback,
we build on deep Q-learning \cite{mnih2013playing} to identify an appropriate 
phase of the action from a provided nominal plan.
Unlike most other applications of 
successful reinforcement learning \cite{mnih2013playing,silver2016mastering},
the biggest challenge is a lack of a robotics simulation software that can generate
realistic haptic signals for a robot to safely
simulate and explore various 
combinations of states with different actions.

%% CONTRIBUTION
To validate our approach, we collect a large number of sequences of haptic feedback 
along with their executed motion
for the task of `turning a knob until it clicks' on objects of various shapes.
We empirically show on a PR2 robot that we can modify a nominal plan 
and successfully accomplish the task using the learned models, 
incorporating tactile sensor feedback on the fingertips of the robot.
In summary, the key contributions of this work are:
\begin{itemize}
    \item an algorithm which learns task relevant representation of haptic feedback 
    \item a framework for modifying a nominal manipulation plan for 
          interactions that involves haptic feedback
    \item an algorithm for learning optimal actions with limited data without simulator
\end{itemize}

%% file: 2_relatedwork.tex
\section{Related Work}
\label{sec:related}

\noindent
\textbf{Haptics.}
Haptic sensors mounted on robots enable 
many different interesting applications.
%at at the tip of the finger allows robot to 
%to classify different material properties. 
Using force and tactile input, a food item
can be classified with characteristics which map to appropriate class of motions 
\cite{gemici2014learning}.
Haptic adjectives such as `sticky' and `bumpy' can be
learned with biomimetic tactile sensors \cite{chu2013using}.
Whole-arm tactile sensing allows fast reaching in dense clutter.
We focus on tasks with a nominal plan (e.g. \cite{sung_robobarista_2015})
but requires incorporating haptic (tactile) sensors 
to modify execution length of each phase of actions.

For closed-loop control of robot,
there is a long history of using different feedback mechanisms 
to correct the behavior \cite{bennett1996brief}.
%When controlling robot or its manipulator with accuracy for complex task,
%some type of automated feedback loop is essential part of control mechanism \cite{bennett1996brief}.
One of the common approaches that involves contact
relies on stiffness control, which uses the pose of an end-effector 
as the error to adjust applied force \cite{salisbury1980active,barry2012manipulation}.
The robot can even self-tune its parameters for its controllers \cite{trimpe2014self}.
%What has also been shown is a robot that can iteratively self-tune its parameters for its controllers \cite{trimpe2014self}.
A robot also uses the error in predicted pose for force trajectories \cite{lenz2015deepmpc}
and use vision for visual servoing \cite{chaumette2006visual}.

Haptic sensors have also been used to provide feedback. % to controller and planner.
A human operator with a haptic interface device can teleoperate a robot remotely \cite{park2006haptic}.
Features extracted from tactile sensors can serve as feedback to planners to slide 
and roll objects \cite{li2013control}.
\cite{pastor2011skill} uses tactile sensor to detect success and failure of manipulation task
to improve its policy.

%%%%%%%%%%%%%%%%%%%%%%%%%%%%%%%%%%%%%%%
\textbf{Partial Observability.}
A POMDP is a framework  
for a robot to plan its actions under uncertainty given that the states are 
often only obtained through noisy sensors \cite{thrun2005probabilistic}.
The framework has been successfully used for many tasks including navigation and grasping \cite{hsiao2007grasping,kurniawati2008sarsop}.
Using wrist force/torque sensors, hierarchical POMDPs help a robot localize certain
points on a table \cite{vien2015touch}.
While for some problems \cite{hsiao2007grasping}, states can be defined
as continuous robot configuration space,
it is unclear what the ideal state space representation is for many complex manipulation tasks.

When the knowledge about the environment or states is not sufficient,
\cite{sallans1999learning} use a fully connected DBN for
learning factored representation online,
while \cite{contardo2014learning} employ a two step method
of first learning optimal decoder then learning to encode.
While many of these work have access to a good environment model, 
or is able to simulate environment 
where it can learn online,
we cannot explore or simulate to learn online.
Also, the reward function is not available.
For training purposes, we perform privileged learning \cite{vapnik2009new}
by providing an expert reward label only during the training phase.

%%%%%%%%%%%%%%%%%%%%%%%%%%%%%%%%%%%%%%%

\textbf{Representation Learning.}
%\todo{DeepLearning}
Deep learning has recently vastly improved the performance of
many related fields such as compute vision (e.g. \cite{krizhevsky2012imagenet})
and speech recognition (e.g. \cite{hannun2014deep}).
In robotics, it has helped robots to 
better classify haptic adjectives by combining images with haptic signals \cite{gao2015deep},
predict traversability from long-range vision \cite{hadsell2008deep},
and classify terrains based on acoustics \cite{valada2015acoustics}.

For controlling robots online,
a deep auto-encoder can learn lower-dimensional embedding from images 
and model-predictive-control (MPC) is used for optimal control \cite{wahlstrom2015pixels}.
%\cite{wahlstrom2015pixels} uses a deep auto-encoder
%to learn lower-dimensional embedding from images 
%and employs model-predictive-control (MPC) for optimal control.
DeepMPC \cite{lenz2015deepmpc} predicts its future
end-effector position with a recurrent network
and computes an appropriate amount of force.
Convolutional neural network 
can be trained to directly map images to motor torques \cite{levine2015end,finn2016deep}.
As mentioned earlier, we only take input of haptic signals,
which suffers from perceptual aliasing,
and contains a lot less information in a single timestep compared to RGB images.

%\todo{Variational}
Recently developed variational Bayesian approach \cite{kingma2013auto,rezende2014stochastic},
combined with a neural network,
introduces a recognition model to approximate intractable true posterior.
Embed-to-Control \cite{watter2015embed} learns embedding from images 
and transition between latent states representing unknown dynamical system.
Deep Kalman Filter \cite{krishnan2015deep} learns very similar temporal
model based on Kalman Filter but is used for counterfactual inference 
on electronic health records.

%%%%%%%%%%%%%%%%%%%%%%%%%%%%%%%%
\begin{figure*}[!ht]
    \vskip -0.15in
    \centering
    %\null \hfill
    \subfloat[\textbf{Graphical Model Rep. of POMDP Model} \label{fig:pomdp}]{%
        % left lower right top
        \includegraphics[width=0.3\textwidth,trim={0cm 0cm 12cm 3cm}, clip]{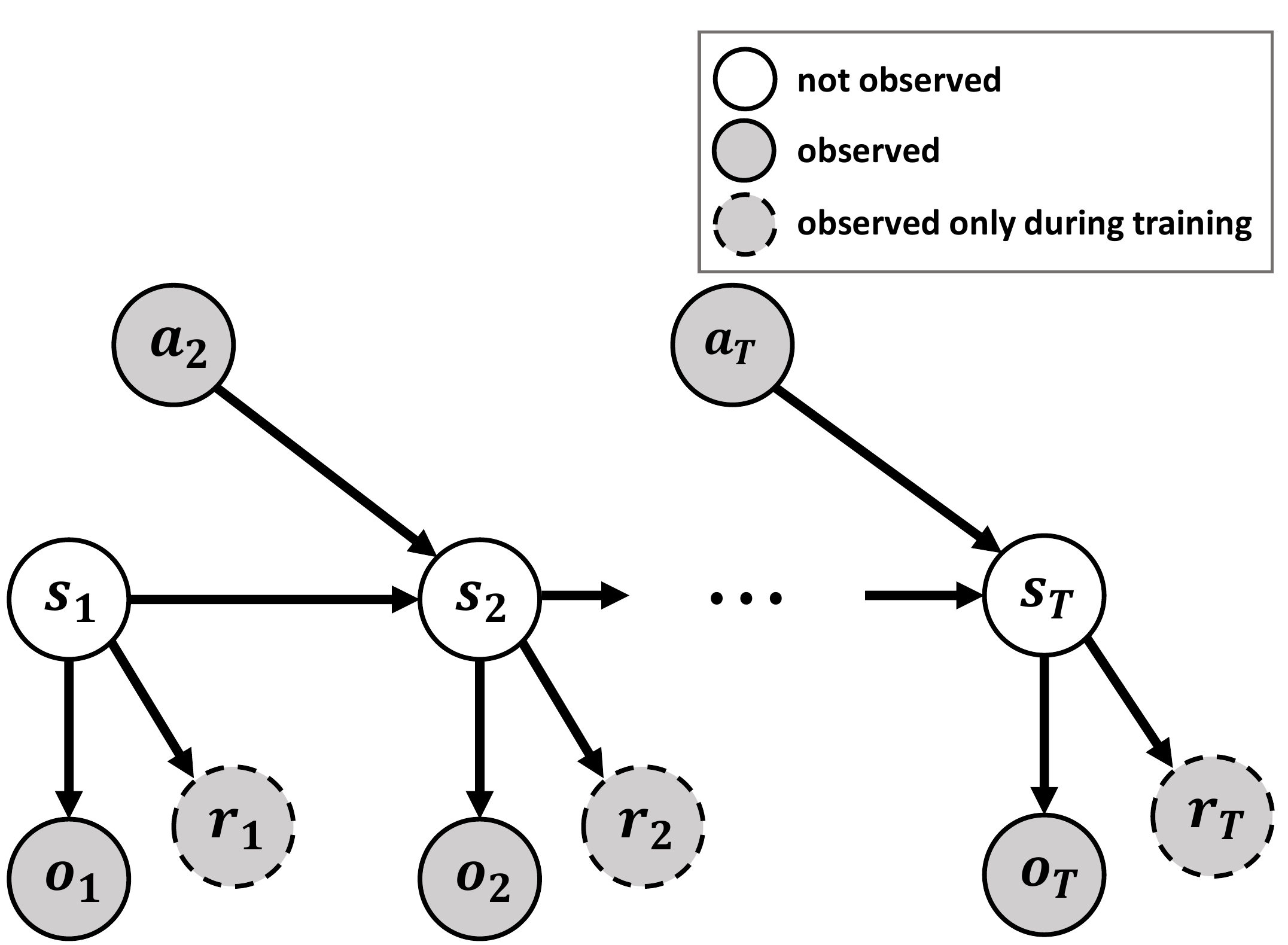}
    }
    \hfill
    \subfloat[\textbf{Tran. Network} \label{fig:trans_net}]{%
        \;\;\;
        \includegraphics[width=0.13\textwidth,trim={0cm 0cm 25cm 14cm}, clip]{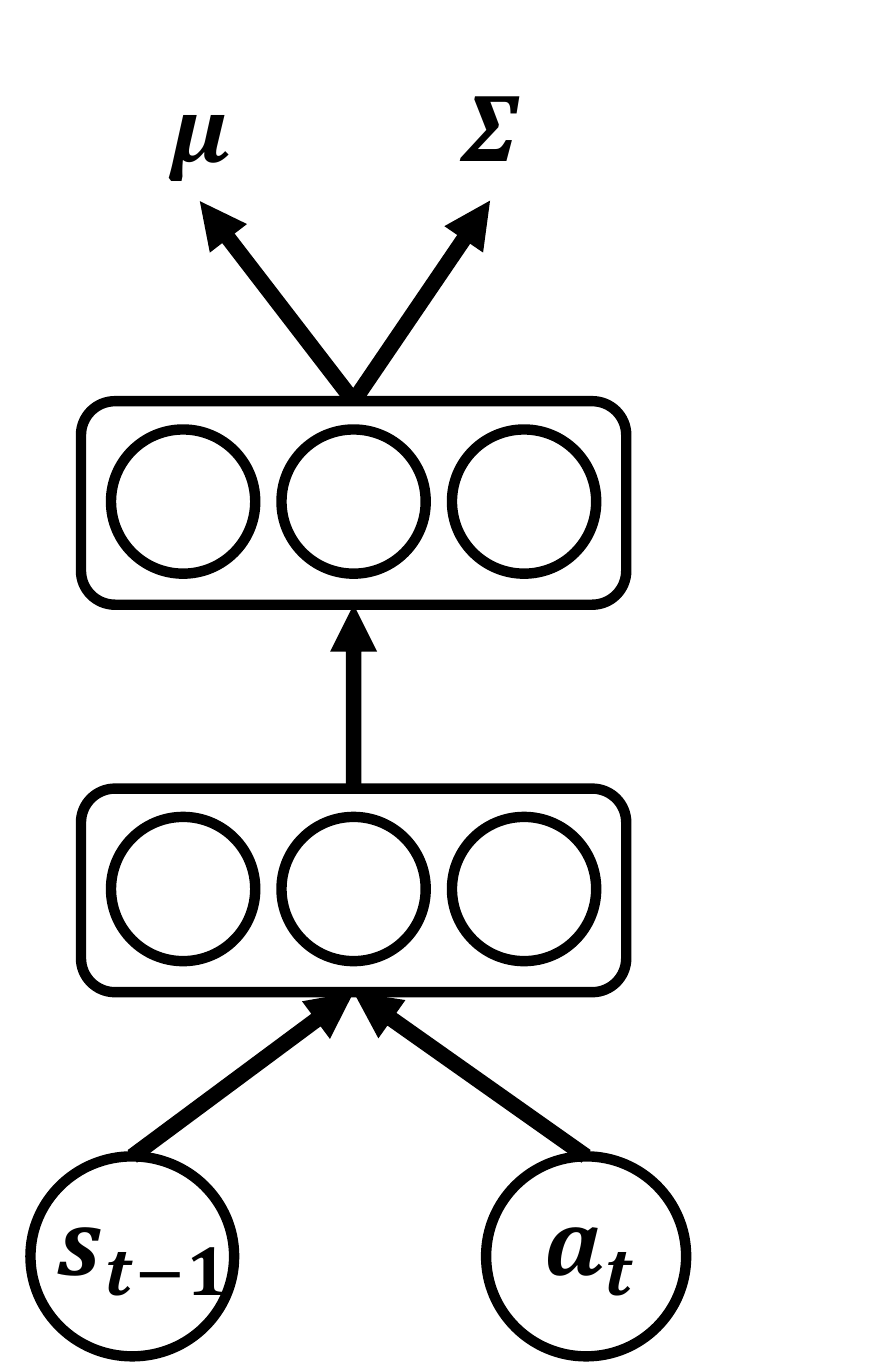}
    }
    \hfill
    \subfloat[\textbf{Deep Recurrent Recognition Network} 
    %for evaluating the posterior. Two LSTM (square boxes) recurrent neural network separately tracks actions and observations.
    \label{fig:var_q}]{%
        \;\;
        \includegraphics[width=0.32\textwidth,trim={0cm 0cm 12cm 11cm}, clip]{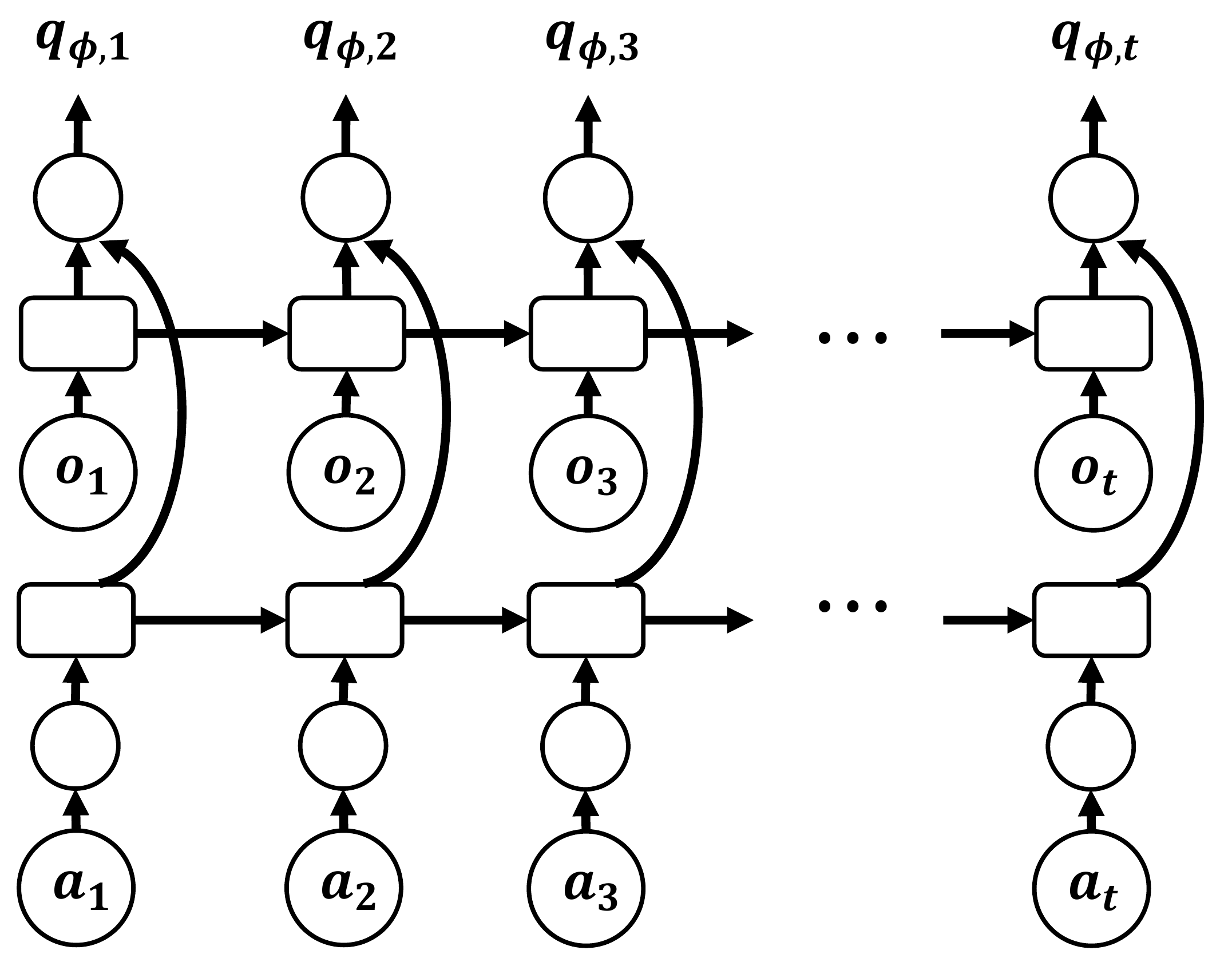}
    }
    \hfill
    \subfloat[\textbf{Deep Q Network} \label{fig:dqn}]{%
        \;\;\;
        \includegraphics[width=0.13\textwidth,trim={0cm 0cm 25cm 14cm}, clip]{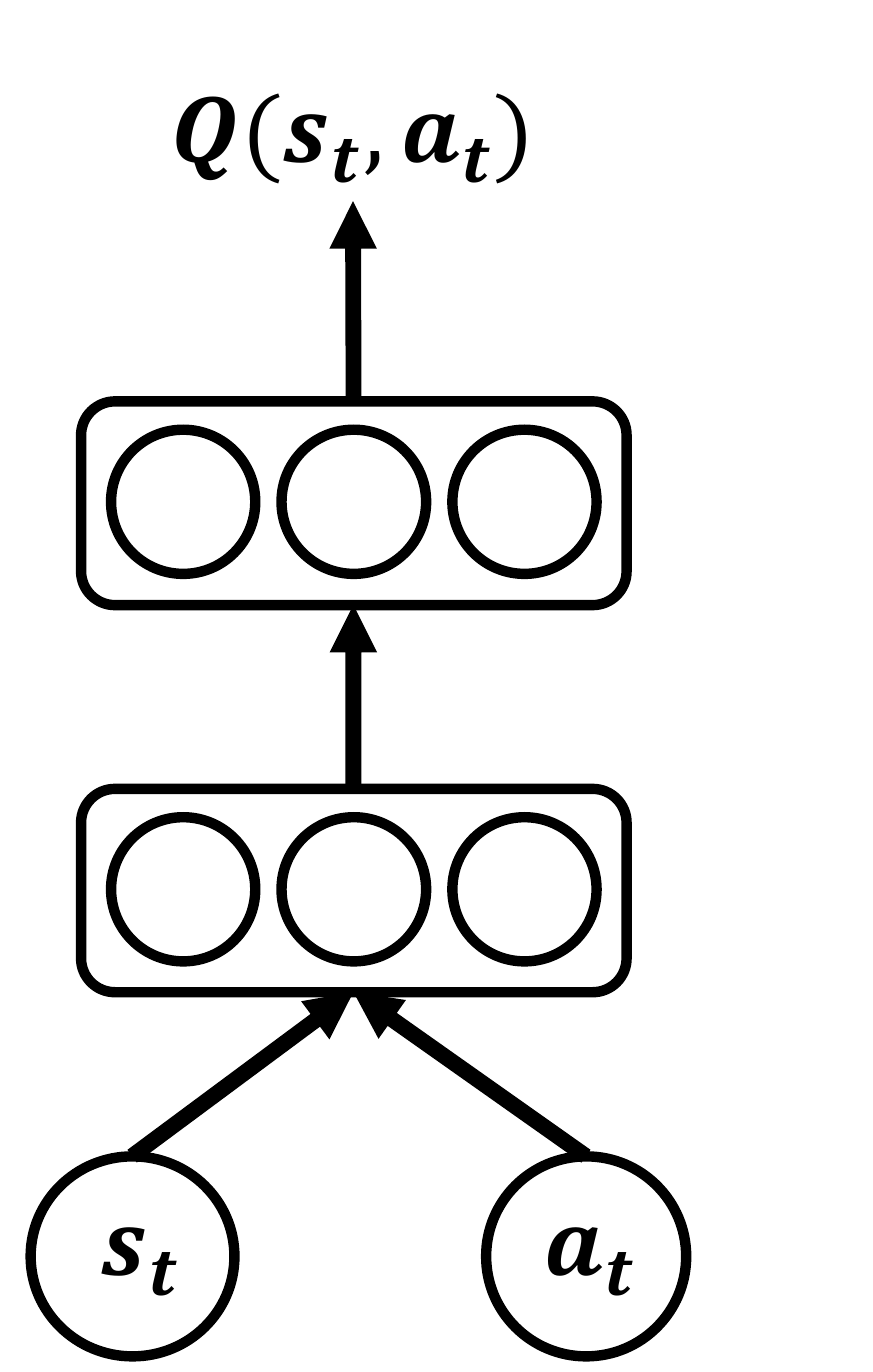}
    }
    %\hfill \null
    \vskip -0.12in
    \caption{
    \textbf{Framework Overview.}
    We model the task that requires incorporation of tactile feedback
    in a partially observable MDP~(a) which its transition and emission functions
    are parametrized by neural networks~(b). 
    To find an appropriate representation of states for the POMDP,
    %To find an appropriate representation that can be used as states for POMDP,
    we approximate the posterior with a Deep Recurrent Recognition Network~(c),
    consisting of two LSTM (square blocks) recurrent networks.
    Deep Q-Network~(d), consisting of two fully connected layers, 
    utilizes a learned representation from (c) and a learned transition model from (a)
    to train Deep Q-Network (d).
    }
    \label{fig:all_models}
\end{figure*}
%%%%%%%%%%%%%%%%%%%%%%%%%%%%%%%%

%\todo{QLearning}
Reinforcement learning (RL), also combined with a neural network,
has recently learned to play computer games by looking at pixels \cite{mnih2013playing,hausknecht2015deep}.
Applying standard RL to a robotic manipulation task, however, is challenging
due to lack of suitable state space representation \cite{finn2016deep}.
Also, most RL techniques rely on trial and error \cite{sutton1998reinforcement}
with the ability to try different actions from different states
and observe reward and state transition.
However,  for many of the robotic manipulation tasks 
that involve physical contact with the environment,
it is too risky  to let an algorithm try different actions,
and reward is not trivial without instrumentation of the environment for many tasks.
In this work,
the robot learns to represent haptic feedback and find optimal control
from limited amount of haptic sequences
despite lack of good robotic simulator for haptic signal.

%% file: 3_approach.tex
%%%%%%%%%%%%%%%%%%%%%%%%%%%%%%%%%%%%%%%%%%%%%%%%%%%
\section{Our Approach}
\label{sec:approach}
\vspace*{\sectionReduceBot}

Our goal is to build a framework
that allows robots to represent and reason about %a manipulation plan
haptic signals generated
by its interaction with an environment.

Imagine you were asked to turn off the hot plate in Figure~\ref{fig:main_fig}
by rotating the knob until it clicks.
In order to do so, you would start by 
rotating the knob clockwise or counterclockwise until it clicks.
If it doesn't click and if you feel the wall,
you would start to rotate it in the opposite direction.
And, in order to confirm that you have successfully completed the task or hit the wall, 
you would use your sense of touch on your finger to feel a click.
There could also be a sound of a click 
as well as other observable consequences,
but you would not feel very confident about the click 
in the absence of haptic feedback.

However, such haptic signal itself does not contain 
sufficient information for a robot to directly act on.
It is unclear what is the best representation for a state of the task,
whether it should only be dependent on states of 
internal mechanisms of the object (which are unknown)
or it should incorporate information about the interaction as well.
The haptic signal is merely a noisy partial observation 
of latent states of the environment, 
influenced by many factors 
such as a type of interaction that is involved
and a type of grasp by the robot.

To learn an appropriate representation of the state,
we first define our manipulation task as 
a POMDP model.
However, posterior inference on such latent state
from haptic feedback is intractable.
In order to approximate the posterior,
we employ variational Bayes methods to jointly
learn model parameters for both a POMDP and an approximate posterior model,
each parametrized
by a deep recurrent neural network.

% dont forget to update that reference!

Another big challenge is the limited opportunity to explore
with different policies to fine-tune the model,
unlike many other applications that employs POMDP or reinforcement learning.
Real physical interactions involving contact are too risky
for both the robot and the environment without 
lots of extra safety measures.
Another common solution is to explore in a simulated environment;
however, none of the available robot simulators, as far as we are aware,
are capable of generating realistic feedback
for objects of our interest.

Instead, we learn offline from previous experiments
by utilizing a learned haptic representation along with its transition model
to explore offline and learn Q-function.

%%%%%%%%%%%%%%%%%%%%%%%%%%%%%
\begin{figure}[tb]
  \vskip -0.05in
  \begin{center}
    % left lower right top
    \includegraphics[width=\columnwidth,trim={0cm 0cm 6.8cm 7.6cm}, clip]{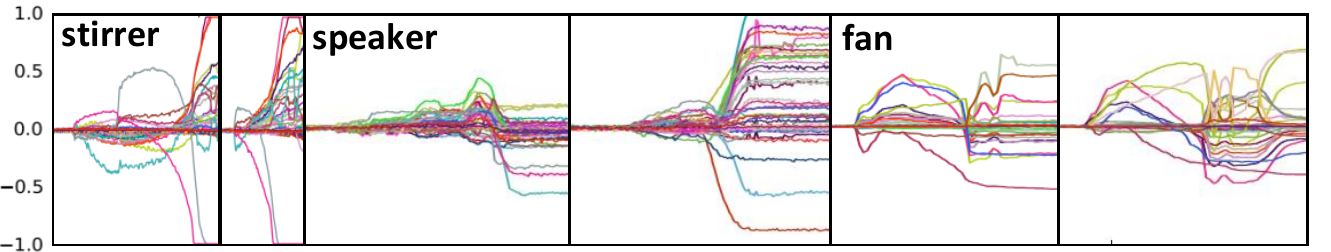}
  \end{center}
  \vskip -0.12in
  \caption{\textbf{Samples of haptic signals} from three different objects
  with a PR2 fingertip tactile sensor. Notice a large variation
  in feedback produced by what humans identify as a `click'.
    }
  \label{fig:haptic_signal_ex}
  \vskip -0.15in
\end{figure}
%%%%%%%%%%%%%%%%%%%%%%%%%%%%%

%%%%%%%%%%%%%%%%%%%%%%%%%%%%%%%%%%%%%%%%%%%%%%%%%%%
\subsection{Problem Formulation}
\label{sec:pf}
\vspace*{\subsectionReduceBot}

Given a sequence of haptic signals ($\vo = o_1, ... , o_t$) up to current time frame $t$
along with a sequence of actions taken ($\va = a_1, ..., a_t$),
our goal is to output a sequence of appropriate state representations ($\vs = s_1, ..., s_t$)
such that we can take an optimal next action $a_{t+1}$ inferred from the current state $s_t$.

%%%%%%%%%%%%%%%%%%%%%%%%%%%%%%%%%%%%%%%%%%%%%%%%%%%
\subsection{Generative Model}
\label{sec:model}
\vspace*{\subsectionReduceBot}

We formulate the task that requires haptic feedback as a POMDP model,
defined as $(S,A,T,R,O)$.
$S$ represents a set of states, $A$ represents a set of actions,
$T$ represents a state transition function,
$R$ represents a reward function, and
$O$ represents an observation probability function.
Fig.~\ref{fig:pomdp} represents a graphical model representation
of a POMDP model and all notations 
are summarized in Table~\ref{tab:notations}.

Among the required definitions of a POMDP model, 
most importantly, 
state $S$ and its representation are unknown.
Thus, all functions $T, R, O$ that rely on states $S$
are also not available.

We assume that all transition and emission probabilities
are distributed as Gaussian distributions; however, they
can take any appropriate distribution for the application.
Mean and variance of each distribution are defined as a function
with input as parent nodes in the graphical model (Fig.~\ref{fig:pomdp}):
\begin{align*}
s_1 &\sim \mathcal{N}(\mathbf{0},\mathbf{I})  \\
s_t &\sim \mathcal{N}(f_{s_{\mu}}(s_{t-1}, a_{t}), f_{s_{\Sigma}}(s_{t-1}, a_{t})^2 \mathbf{I})  \\
o_t &\sim \mathcal{N}(f_{o_{\mu}}(s_{t}), f_{o_{\Sigma}}(s_{t})^2 \mathbf{I})  \\
r_t &\sim \mathcal{N}(f_{r_{\mu}}(s_{t}), f_{r_{\Sigma}}(s_{t})^2 \mathbf{I}) 
\end{align*}
We parametrize each of these functions as a neural network.
Fig.~\ref{fig:trans_net} shows a two layer network for parametrization of the transition function, 
and emission networks take a similar structure.
The parameters of these networks form the parameters of the generative model
$\theta = \{s_{\mu},s_{\Sigma},o_{\mu},o_{\Sigma},r_{\mu},r_{\Sigma} \}$.

%%%%%%%%%%%%%%%%%%%%%%%%%%%%%%%%%%%%%%%%%%%%%%%%%%%
\begin{table}[tb]
%\vskip -.1in
\begin{center}
\caption{
Summary of Notations.
}
\vskip -.05in
\begin{tabular}{@{}r|l@{}}
\hline
Notations & Descriptions \\ 
\hline
$S$ & continuous state space (a learned representation) \\
$O$ & observation probability  $(S \rightarrow O)$ of haptic signal \\
$T$ & conditional probability between states $(S \times A \rightarrow S)$ \\
$A$ & a set of possible actions to be taken at each time step \\
$R$ & a reward function $(S \rightarrow \mathbb{R})$ \\ 
\hline
$p_\theta$ & a generative model for $O$ and $R$  \\ 
$\theta$ & parameters of generative model \\
$q_\phi$ & an approximate posterior distribution \\
& \;(a recognition network for representing haptic signal) \\ 
$\phi$ & parameters of recognition network (recurrent neural network)  \\  
\hline
$Q(s,a)$ & an approximate action-value function $(S \times A \rightarrow \mathbb{R})$  \\ 
$\gamma$ & a discount factor \\ 
\hline
\end{tabular}
\label{tab:notations}
\end{center}
\vskip -.15in
\end{table}
%%%%%%%%%%%%%%%%%%%%%%%%%%%%%%%%%%%%%%%%%%%%%%%%%%%

%%%%%%%%%%%%%%%%%%%%%%%%%%%%%%%%%%%%%%%%%%%%%%%%%%%
\subsection{Deep Recurrent Recognition Network}
\label{sec:recognition}
\vspace*{\subsectionReduceBot}

Due to non-linearity of multi-layer neural network,
computing the posterior distribution $p(\vs|\vo,\vr,\va)$
becomes intractable \cite{krishnan2015deep}.
%To address the intractability of computing the posterior
%distribution $p(\vs|\vo,\vr,\va)$,
The variational Bayes method \cite{kingma2013auto,rezende2014stochastic}
allows us to approximate the 
real posterior distribution with a recognition network (encoder) $q_{\phi}(\vs|\vo,\vr,\va)$.

Although it is possible to build a recognition network $q_{\phi}(\vs|\vo,\vr,\va)$ 
that takes the reward $\vr$ as a part of the input,
such recognition network would not be useful during a test time
when the reward $\vr$ is not available.
Since a reward is not readily available for many of the interaction tasks,
we assume that the sequence of rewards $\vr$ is available only during 
a training phase given by a expert.
Thus, we build an encoder $q_{\phi}(\vs|\vo,\va)$ without a reward vector
while our goal will be to reconstruct a reward $\vr$ as well (Sec.~\ref{sec:variational}).

Among many forms and structures $q_{\phi}$ could take,
through validation with our dataset,
we chose to define $q_{\phi,t}(s_t|o_1,...,o_t,a_1,...,o_t)$
as a deep recurrent network with two long short-term memory (LSTM) layers 
as shown in Fig.~\ref{fig:var_q}.

%%%%%%%%%%%%%%%%%%%%%%%%%%%%%%%%%%%%%%%%%%%%%%%%%%%

%%%%%%% Variational %%%%%%
\input{3_variational}

%%%%%%% CONTROL %%%%%%

\input{3_control}

%% file: 3_variational.tex
\subsection{Maximizing Variational Lower-bound}
\label{sec:variational}
\vspace*{\subsectionReduceBot}

To jointly learn parameters for the generative $\theta$ and the recognition network $\phi$,
our objective is to maximize likelihood of the data: 
$$max_\theta \big[ log \;p_\theta (\vo,\vr|\va) \big]$$
Using a variational method, a lower bound on conditional log-likelihood is defined as:
\begin{align*}
log \; p_\theta (\vo,\vr|\va) &= D_{KL}(q_\phi(\vs|\vo,\vr,\va) ||p_\theta(\vs|\vo)) + \mathcal{L}(\theta,\phi) \\
  &\geq  \mathcal{L}(\theta,\phi) 
%log \; p_\theta (\vo,\vr|\va) &\geq  \mathcal{L}(\theta,\phi) \\
\end{align*}
Thus, to maximize $max_\theta \big[ log \; p_\theta (\vo,\vr|\va) \big]$,
the lower bound $\mathcal{L}(\theta,\phi)$ can instead be maximized.
\begin{align}
\mathcal{L}(\theta,\phi) &= -D_{KL}\big( q_\phi(\vs|\vo,\vr,\va) || p_\theta(\vs|\va) \big)  \notag \\
&\quad \; + \mathbb{E}_{q_\phi(\vs|\vo,\vr,\va)} \big[ log \; p_\theta(\vo,\vr|\vs,\va) \big]  \label{eq:recon}
\end{align}

Using a reparameterization trick \cite{kingma2013auto} twice, 
we arrive at following lower bound 
(refer to Appendix for full derivation): 
%(refer to Sec.~\ref{sec:derivation} for full derviation): 
\begin{align}
&\mathcal{L}(\theta,\phi) \approx -D_{KL}\big( q_\phi(s_1|\vo,\vr,\va) || p(s_1) \big) \notag \\
& \quad - \frac{1}{L} \sum_{t=2}^{T} \sum_{l=1}^{L} \big[ 
D_{KL}\big(q_\phi(s_t|s_{t-1},\vo,\vr,\va)||p(s_t|s_{t-1}^{(l)},u_{t-1})\big) \big] \notag \\
& \quad + \frac{1}{L} \sum_{l=1}^{L} \big[ log \; p_\theta(\vo|\vs^{(l)}) + log \; p_\theta(\vr|\vs^{(l)}) \big] \notag \\
& \;\;\;\; \text{ where } \vs^{(l)}=g_\phi(\epsilon^{(l)}, \vo, \vr, \va) \text{ and } \epsilon^{(l)} \sim p(\epsilon)  \label{eq:recon_final}
\end{align}
We jointly back-propagate on neural networks for both sets of encoder $\phi$ and decoder $\theta$ parameters 
with mini-batches to maximize the lower bound using AdaDelta \cite{zeiler2012adadelta}.

%% file: 3_control.tex
%%%%%%%%%%%%%%%%%%%%%%%%%%%%%%%%%%%%%%%%%%%%%%%
%%%%%%%%%%%%%%% DEEP Q LEARNING %%%%%%%%%%%%%%%
\begin{algorithm}[tb]
\caption{Deep Q-Learning in Learned Latent State Space} 
\label{algo:deep_q}
\begin{algorithmic}
\small
\State $D_{gt} = \{\}$
\Comment{``ground-truth'' transitions by $q_\phi$}
\ForAll{timestep $t$ of $(\vo, \va)$ in training data $(i)$} 
    \State $s_t, s_{t+1} \gets $ $q_{\phi,\mu} + \epsilon \; q_{\phi,\Sigma}$ where $\epsilon \sim p(\epsilon)$
    \State $D_{gt} \gets D_{gt} \cup \langle s_t^{(i)},  a_{t+1}^{(i)}, r_{t+1}^{(i)},  s_{t+1}^{(i)} \rangle $
\EndFor
\Loop
    \State $D_{explore} = \{\}$ 
    \Comment{explore with learned transition}
    \ForAll{$s_t^{(i)}$ in training data that succeeded} 
        %\State $a_{t} \gets \argmax_{a \in A} Q(s,a)$ 
        \State $a_{t+1} = \left\{\begin{array}{ll}
                rand(a \in A)                  & \text{with prob. $\epsilon$} \\
                \argmax_{a \in A} Q(s_t^{(i)},a)         & \text{otherwise}
           \end{array}\right.$
        \State $r_{t+1} = \left\{\begin{array}{ll}
                r_t^{(i)} & \text{if $a_{t+1} == a_{t+1}^{(i)}$} \\
                -1        & \text{otherwise}
           \end{array}\right.$
        \State $s_{t+1} \gets T(s_t^{(i)},a_t)$
        \State $D_{explore} \gets D_{explore} \cup \langle s_t^{(i)},  a_{t+1}, r_{t+1}, s_{t+1} \rangle $
    \EndFor
    
    \State $D \gets D_{gt} \cup D_{explore}$ 
    \Comment{update deep Q-network}
    \ForAll{minibatch from $D$}
        \State $y_{t} \gets r_{t} + \gamma max_{a'} Q(s_{t+1},a')$
        \State Take gradient with loss $[y_{t} - Q(s_{t}, a_{t+1})]^2$
    \EndFor
\EndLoop
\end{algorithmic}
%\vskip -.2in
\end{algorithm}

%%%%%%%%%%%%%%%%%%%%%%%%%%%%%%%%%%%%%%%%%%%%%%%

\subsection{Optimal Control in Learned Latent State Space}
\label{sec:dql}
\vspace*{\subsectionReduceBot}

After learning a generative model for the POMDP and a recognition network
using a variational Bayes method,
we need an algorithm for making an optimal decision in learned representation 
of haptic feedback and action.
We employ a reinforcement learning method, Q-Learning,
which learns to approximate an optimal action-value function \cite{sutton1998reinforcement}.
The algorithm computes a score for each
state action pair: $$Q:S \times A \rightarrow \mathbb{R}$$
The $Q$ function is approximated by a two layer neural network as shown in Fig.~\ref{fig:dqn}.

In a standard reinforcement learning setting,
in each state $s_t$, an agent learns by exploring the selected action $\argmax_{a \in A} Q(s_t,a)$
with a current $Q$ function.
However, doing so requires an ability to actually take or simulate the chosen action 
from $s_t$ and observe $r_{t+1}$ and $s_{t+1}$.
However, there does not exist a good robotics simulation software
that can simulate complex interactions between a robot and an object
and generate different haptic signals.
Thus, we cannot freely explore any states.

Instead, we first take all state transitions and rewards 
$\langle s_t^{(i)}, a_{t+1}^{(i)}, r_{t+1}^{(i)}, s_{t+1}^{(i)} \rangle$
from the $i$-th training data sequence and store in $D_{gt}$.
Both $s_t^{(i)}$ and $s_{t+1}^{(i)}$ are computed by the recognition network $q_\phi$
with a reparameterization technique (similar to Sec.~\ref{sec:variational}).

At each iteration, we first have an exploration stage.
For explorations, we start from states $s_t^{(i)}$ of training sequences 
that  resulted in successful completion of the task
and choose an action $a_{t+1}$ with $\epsilon$-greedy.
With the learned transition function $T$ (Sec.~\ref{sec:model}),
the selected action $a_{t+1}$ is executed from $s_t^{(i)}$.
However, since we are using a learned transition function, 
any deviation from the distribution of training
data could result in unexpected state,
unlike explorations in a real or a simulated environment.
%unlike explorations in real environment or simulated environment.

Thus, if the optimal action $a_{t+1}$ using a current Q-function deviates
from the ground-truth action $a_{t+1}^{(i)}$, the action is penalized with a negative reward
to prevent deviations into unexplored states.
%We take advantage of the learned POMDP model earlier
%to do minimal exploration.
If the optimal action is same as the ground-truth,
the same reward as the original is given.
For such cases,
the only difference from the ground-truth would be in $s_{t+1}$,
which is inferred by the learned transition function.
%For such cases, %cases where optimal action is same as groundtruth,
%the only difference would be $s_{t+1}$ which in this case computed by 
%learned transition function.
All exploration steps are recorded in $D_{explore}$.

After the exploration step in each iteration,
we take minibatches from $D = D_{gt} \cup D_{explore}$
and backpropagate on the deep Q-network with the loss function:
$$[r_t + \gamma max_{a'} Q(s_{t+1},a')] - Q(s_{t}, a_{t})]^2$$
The algorithm is summarized in Algorithm~\ref{algo:deep_q}.

%% file: 4_results.tex
%%%%%%%%%%%%%%%%%%%%%%%%%%%%%
\begin{figure}[tb]
\vskip -0.1in
  \begin{center}
    \includegraphics[width=\columnwidth,height=6cm,trim={0cm 0cm 2.5cm 5cm}, clip]{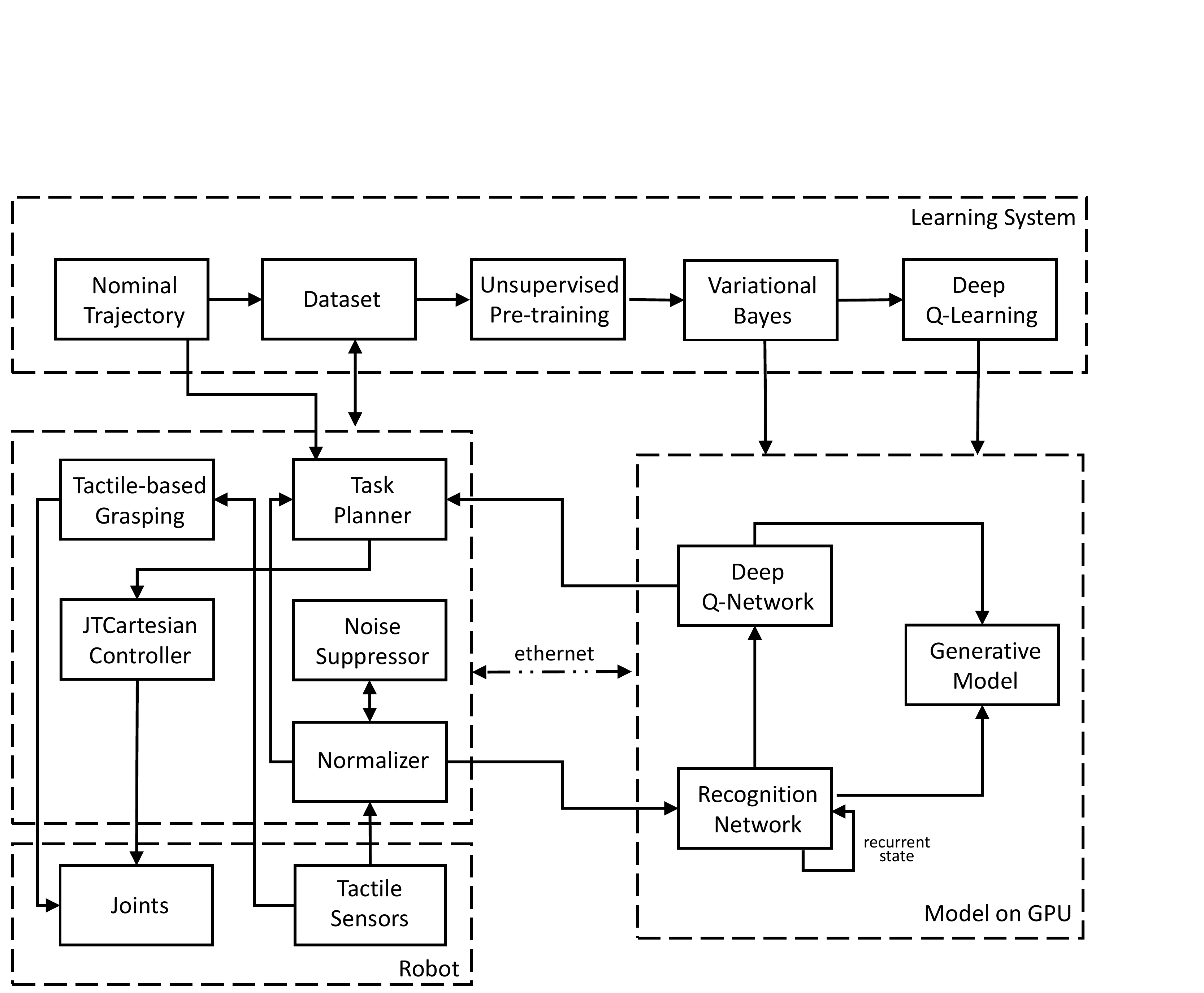}
  \end{center}
  \vskip -0.15in
  \caption{
    \textbf{System Details} of our system for learning and robotic experiments. 
    }
  \label{fig:system}
  \vskip -0.15in
\end{figure}
%%%%%%%%%%%%%%%%%%%%%%%%%%%%%

%%%%%%%%%%%%%%%%%%%%%%%%%%%%%%%%%
\input{4_system_details}

%%%%%%%%%%%%%%%%%%%%%%%%%%%%%%%%%

%%%%%%%%%%%%%%%%%%%%%%%%%%%%%
\begin{figure}[tb]
\vskip -0.1in
  \begin{center}
    \includegraphics[width=\columnwidth,trim={0cm 16cm 2cm 0cm}, clip]{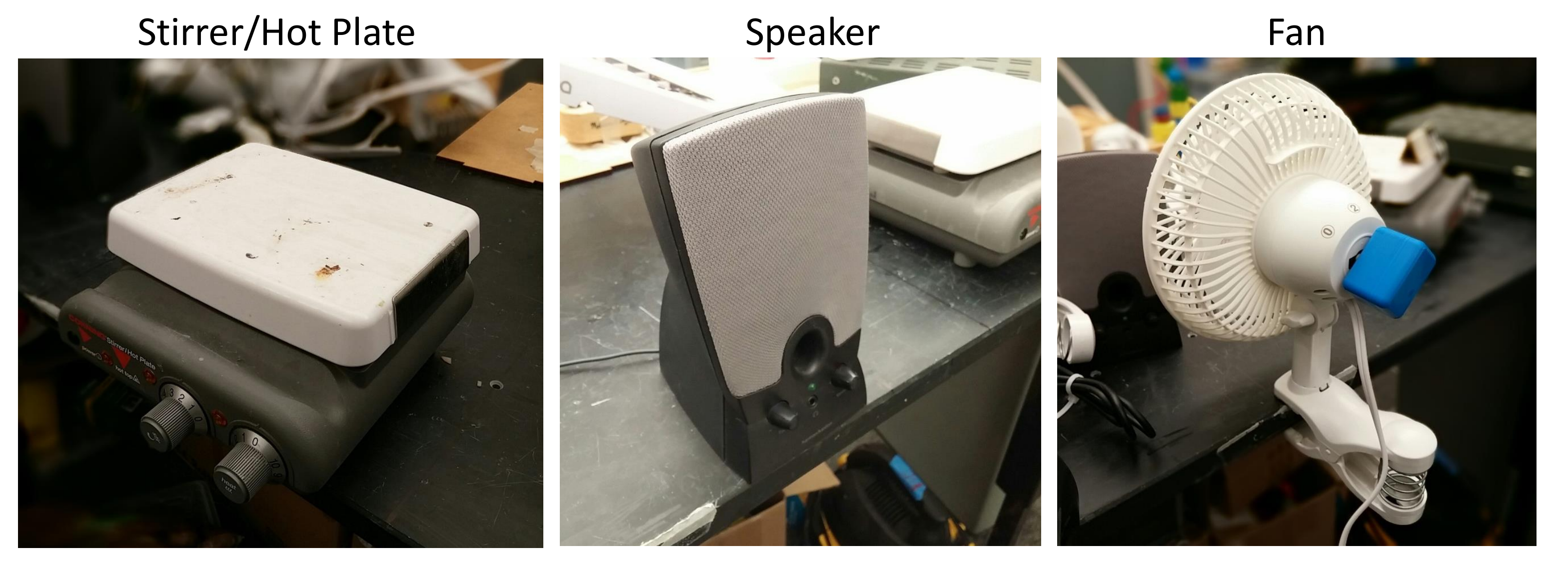}
  \end{center}
  \vskip -0.2in
  \caption{
    A set of objects used for experiment.  All three objects
    have different surface area and shape, which results in vastly
    different types of \textit{`clicks'} when observed via a tactile sensor.
    }
  \label{fig:experiment_obj}
  \vskip -0.17in
\end{figure}
%%%%%%%%%%%%%%%%%%%%%%%%%%%%%

%%%%%%%%%%%%%%%%%%
\input{4_table}

%%%%%%%%%%%%%%%%%%

\section{Experiments \& Results}
\label{sec:experiments}
\vspace*{\sectionReduceBot}

In order to validate our approach, we perform a series of experiments
on our dataset and on a PR2 robot.

%%%%%%%%%%%%%%%%%%%%%%%%%%%%%%%%%
\input{4_results_dataset}

%%%%%%%%%%%%%%%%%%%%%%%%%%%%%%%%%

%%%%%%%%%%%%%%%%%%%%%%%%%%%%%%%%%
\input{4_results_baseline}
%%%%%%%%%%%%%%%%%%%%%%%%%%%%%%%%%

%%%%%%%%%%%%%%%%%%%%%%%%%%%%%%%%%%%%%%%
\subsection{Results and Discussion}
\vspace*{\subsectionReduceBot}

To evaluate all models,
we perform two types of experiments --- haptic signal prediction and robotic experiment.

\noindent
\textbf{Haptic Signal Prediction.}
We first compare our model against baselines %on our dataset
on a task of predicting future haptic signal.
%We first compare our model against baselines on our dataset
%on a task of predicting future haptic signal.
%From the dataset, 
For all sequences that 
either eventually succeeded or failed,
%successful or successfully executed or failed,
we take every timestep $t$,
%from the sequence,
and predict timestep $t+1$ $(0.05secs)$, $t+5$ $(0.25secs)$ and $t+10$ $(0.5secs)$.
The prediction is made by encoding (recognition network)
a sequence up to time $t$ and then transiting encoded states with a learned transition model 
to the future frames of interest.
We take the L2-norm of the prediction of $44$ sensor values (which are in newtons) and take the average of that result.
The result is shown in the middle column of Table~\ref{tab:results}.

\noindent
\textbf{Robotic Experiment.}
On a PR2 robot, we test the task of turning a knob
until it clicks on three different objects: stirrer, speaker, and desk fan (Fig.~\ref{fig:experiment_obj}).
The right hand side of Table~\ref{tab:results} shows the result of over $200$ executions.
%on a physical robot.
Each algorithm was tested on each object at least $15$ times.

\noindent
\textbf{Can it predict future haptic signals?}
When it predicts randomly (\textit{chance}), regardless of the 
timestep, it has an average of $6.7$. 
When the primary goal is to be able to
perform the next haptic signal prediction, for one step prediction,
recurrent-network as representation baseline 
performs best of $0.330$ among all models, while ours performed $0.718$.
%On the other hand, our model performed at $0.718$ which
%is much larger.
On the other hand, our model does not diverge and performs consistently well.
%However, very interestingly, our model does not diverge and performs consistently well.
After $0.5 secs$, when other models started to diverge to an error of $1.757$ or much larger,
our model still had prediction error of $0.782$.

\begin{figure*}[tb]
  \vskip -0.05in
  \begin{center}
    \includegraphics[width=\textwidth,trim={0cm 0cm 0cm 0cm}]{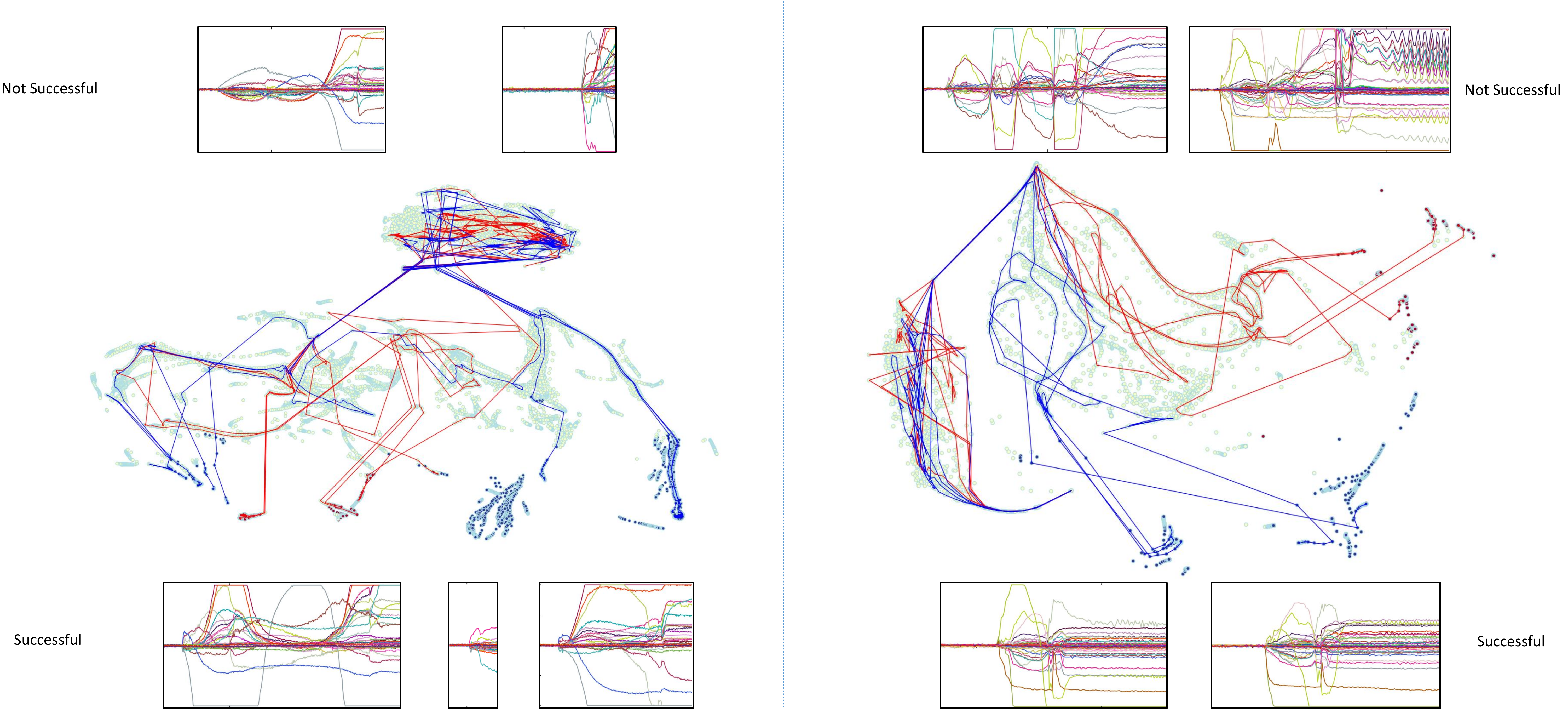}
  \end{center}
  \vskip -0.15in
  \caption{
    \textbf{Projection of learned representation} of haptic feedback
    using t-SNE \cite{tsne} for `stirrer' and `fan'.
    Each dot represents an inferred state at each time frame, and
    blue and red dots represents positive and negative reward at those time frame.
    Here we show some of successful (blue) and unsuccessful (red) sequences.
    For both objects, notice both classes initially starts from similar state and then
    diverges, forming clusters. Several successful and unsuccessful haptic signals
    are shown as well.
    }
  \label{fig:proj}
  \vskip -0.1in
\end{figure*}

\noindent
\textbf{What does learned representation represent?}
We visualize our learned embedding space of haptic feedback
using t-SNE \cite{tsne} in Fig.~\ref{fig:proj}.
Initially, both successful (blue paths) and unsuccessful (red paths) all starts from similar states 
but they quickly diverge into different clusters of paths
much before they eventually arrive at states that were
given positive or negative rewards shown as blue and red dots.

%%%%%%%%%%%%%%%%%%%%%%%%%%%%%
%%%%%%%%%%%%%%%%%%%%%%%%%%%%%

%%%%%%%%%%%%%%%%%%%%%%%%%%%%%%%%%%%%%%%
%\subsection{Robotic Experiment}
%\label{sec:robotic_exp}

\noindent
\textbf{Does good representation lead to successful execution?}
%For all three objects, our robot is able to successfully
%execute successful tasks $80.0\%$, $73.3\%$, and $86.7\%$
%respectively, performing the highest compared to any other model.
Our model allows robot to successfully
execute on the three objects $80.0\%$, $73.3\%$, and $86.7\%$ respectively, 
performing the highest compared to any other models.
The next best model which uses recurrent network as representation
performed at $63.2\%, 68.4\%,$ and $70.0\%$.
%The other best model which uses recurrent network as representation
%performed at $63.2\%, 68.4\%,$ and $70.0\%$.
However, note that this baseline still take advantage of our 
Q-learning method.
Our model that did not take advantage of 
simulated exploration performed much poorly ($35.0\%, 33.3\%,$ and $52.6\%$),
showing that good representation combined with our Q-learning method 
leads to successful execution of the tasks.

\noindent
\textbf{Is recurrent network necessary for haptic signals?} % really different from images?}
\emph{Non-recurrent recognition network} 
quickly diverged to extremely large number of $3.2e7$ 
even though it successfully predicted $1.389$ for a single step prediction.
Note that it takes windowed haptic sequence of last $5$ frames as input.
Unlike images, short window of data does not hold enough information 
about haptic sequence which lasts much longer timeframe.
For robotic experiment, non-recurrent network performed $52.9\%, 57.9\%,$ and $62.5\%$
even with our Q learning method.

\noindent
\textbf{How accurately does it perform the task?}
When our full model was being tested on three objects,
we also had one of the author observe (visually and audibly) very closely 
and press a button as soon as a click occurs.
On successful execution of the task, 
we measure the time difference between the time the robot
stops turning and the time the expert presses the key, and the results are shown in Table~\ref{tab:time}.

\iffalse
\begin{table}[h!]
\begin{center}
\begin{tabular}{@{}c|c|c@{}}
\hline
\emph{Stirrer} & \emph{Speaker} & \emph{Desk Fan} \\ \hline
$0.180$ secs $(\pm 0.616)$ & $0.539$ secs $(\pm 1.473)$ & $-0.405$ secs $(\pm 0.343)$ \\ \hline  
\end{tabular}
\end{center}
\end{table}
\fi

%%%%%%%%%%%%%%%%%%%%%%%%%%%%%%%%%%%%%%%%
\begin{table}[tb]
\vskip -.05in
\begin{center}
\caption{
\textbf{Time difference} between the time the robot stopped 
and the time the expert indicated it `clicked'.
}
\vskip -.05in
\begin{tabular}{@{}c|c|c@{}}
\hline
\emph{Stirrer} & \emph{Speaker} & \emph{Desk Fan} \\ \hline
$0.180$ secs $(\pm 0.616)$ & $0.539$ secs $(\pm 1.473)$ & $-0.405$ secs $(\pm 0.343)$ \\ \hline  
\end{tabular}
\label{tab:time}
\end{center}
\vskip -.2in
\end{table}
%%%%%%%%%%%%%%%%%%%%%%%

The positive number represents that the model was delayed than the expert
and the negative number represents that the model transitioned earlier.
%As the numbers suggest, 
Our model only differed from human
with an average of $0.37$ seconds.
All executions of tasks were performed at same translational and rotational velocity
as the data collection process.

Note that just like a robot has a reaction time to act on perceived feedback, %knowledge,
an expert has a reaction time to press the key.
However, since the robot was relying on haptic feedback
while the observer was using every possible human senses available
including observation of the consequences without touch,
some differences are expected.
We also noticed that fan especially had a delay in visible consequences
compared to the haptic feedback because robot was rotating these knobs slower
than normal humans would turn in daily life;
thus, the robot was able to react $0.4$ seconds faster.

Video of robotic experiments are available at this website:\\
\url{http://jaeyongsung.com/haptic_feedback/}

%% file: 4_system_details.tex
%%%%%%%%%%%%%%%%%%%%%%%%%%%%%%%%%

\section{System Details}
\label{sec:system}
\vspace*{\sectionReduceBot}

\textbf{Robotic Platform.} 
All experiments were performed on a PR2 robot,
a mobile robot with two 7 degree-of-freedom arms.
Each two-fingered end-effector has an array of tactile sensors located at its tips.
We used a Jacobian-transpose based
JTCartesian controller \cite{jtcartesian}
for controlling its arm during experiments.

For stable grasping,
we take advantage of the tactile sensors
to grasp an object.
The gripper is slowly closed until certain thresholds
are reached on both sides of the sensors,
%The gripper is closed slowly until both ends of the gripper
%feel certain threshold of force in both sensors,
allowing the robot to easily adapt to 
objects of different sizes and shapes.
To avoid saturating the tactile sensors,
%and prevent crushing objects,
the robot does not grasp the object with maximal force.

\textbf{Tactile Sensor.}
Each side of the fingertip of a PR2 robot is equipped with RoboTouch tactile sensor,
an array of 22 tactile sensors covered by protective silicone rubber cover.
The sensors are designed to detect range of $\numrange{0}{30}$ psi ($\numrange{0}{205}$ kPa)
with sensitivity of $0.1$ psi ($0.7$ kPa)
at the rate of $35$ Hz.

We observed that each of the $44$ sensors 
has a significant variation and noise in raw sensor readings with drifts over time.
To handle such noise, values are first offset by starting values when interaction 
between an object and the robot started (\emph{i.e.} when a grasp occurred).
Given the relative signals, we find a normalization
value for each of $44$ sensors such that none of the values goes 
above $0.05$ when stationary and all data is clipped to the range of $-1$ and $1$.
Normalization takes place by recording few seconds of sensor readings after grasping.
%without moving any part of the robot.

\textbf{Learning Systems.}
For fast computation and executions,
we offload all of our models onto a remote 
workstation with a GPU connected over a direct ethernet connection.
Our models run on a graphics card using Theano \cite{Bastien-Theano-2012},
and our high level task planner
sends a new goal location at the rate of $20$ Hz.
The overall system detail is shown in Figure~\ref{fig:system}.

%% file: 4_table.tex
%%%%%%%%%%%%%%%%%%%%%%%%%%%%%%%%%%%%%%%%
\begin{table*}[tb]
\vskip -.05in
\begin{center}
\caption{
\textbf{Result of haptic signal prediction and robotic experiment.}
The prediction experiment reports the average L2-norm from the haptic signal (44 signals in newtons)
and the robotic experiment reports the success rate. It shows the results of more than 
200 robotic experiments.
}
\vskip -.1in
\begin{tabular}{r|c|c|c|c|c|c}
%\begin{tabular}{@{}r|c|c|c|c|c|c@{}}
\hline
 & \multicolumn{3}{c|}{\textbf{Haptics Prediction}} & \multicolumn{3}{c}{\textbf{Robotic Experiment}} \\ \hline
 &  \emph{$0.05 secs$} & \emph{$0.25 secs$} & \emph{$0.50 secs$} &  \emph{Stirrer} & \emph{Speaker} & \emph{Desk Fan} \\ \hline 
\emph{Chance}  & $6.68$ ($\pm 0.18$) & $6.68$ ($\pm 0.17$) & $6.69$ ($\pm 0.18$)  
& $31.6\%$  & $38.1\%$ & $28.5\%$  \\ \hline
\emph{Non-recurrent Recognition Network \cite{watter2015embed}}  
& $1.39$ ($\pm 2.51$) & $5.03e5$ ($\pm 5.27e7$) & $3.23e7$ ($\pm 1.07e10$) 
& $52.9\%$ & $57.9\%$ & $62.5\%$ \\ \hline
\emph{Recurrent-network as Representation  \cite{lenz2015deepmpc}} & $\textbf{0.33}$ ($\pm 0.01$) & $1.01$ ($\pm 0.09$) & $1.76$ ($\pm 0.03$) 
& $63.2\%$ & $68.4\%$ & $70.0\%$ \\ \hline
\emph{Our Model without Exploration}  & - & - & -                      
 & $35.0\%$ & $33.3\%$ & $52.6\%$ \\  \hline
\textbf{\emph{Our Model}}        & $0.72$ ($\pm 0.08$) & $\textbf{0.79}$ ($\pm 0.09$) & $\textbf{0.78}$ ($\pm 0.10$)   
& $\textbf{80.0}\%$ & $\textbf{73.3}\%$ & $\textbf{86.7}\%$ \\ \hline  
\end{tabular}
\label{tab:results}
\end{center}
\vskip -.15in
\end{table*}
%%%%%%%%%%%%%%%%%%%%%%%%%%%%%%%%%%%%%%%

%% file: 4_results_dataset.tex
%%%%%%%%%%%%%%%%%%%%%%%%%%%%%%%%%%%%
\subsection{Dataset}
\vspace*{\subsectionReduceBot}

In order to test our algorithm that learns to represent haptic feedback,
we collected a dataset of three different objects --- a stirrer, a speaker, 
and a desk fan (Fig.~\ref{fig:experiment_obj}) --- each of which 
have a knob with a detent structure 
(an example CAD model shown in Fig.~\ref{fig:main_fig}).
Although these objects internally have some type of a detent structure 
that produce a feedback that humans would identify as a `click',
each `click' from each object is very distinguishable. % to humans.
%Not only these produce different feedback,
%Different shapes of objects and the flat surface of the two fingers
%result in vastly differently tactile sensor readings as shown in Fig.~\ref{fig:haptic_signal_ex}.
As shown in Fig.~\ref{fig:haptic_signal_ex},
different shapes of objects and the flat surface of the two fingers
result in vastly differently tactile sensor readings.

%%%%%%%%%%%%% CONNECT TO OUR MODEL %%%%%%%%%%%%%%%%%
In our model, 
for the haptic signals $\vo$, we use a vector of normalized $44$ tactile sensor array
as described in Sec.~\ref{sec:system}.
The reward $\vr$ is given as one of three classes at each time step,
representing a positive, a negative and a neutral reward.
For every object, action $\va$ is 
an array of binary variables, each representing a phase in its nominal plan.

%%%%%%%%%%%%% OBJECTS %%%%%%%%%%%%%%
In more detail, 
the \textbf{stirrer} (hot plate)
has a knob with a diameter of $22.7 mm$ with a depth of $18.7 mm$,
%and the haptic feedback requires and lasts
which produces a haptic feedback that lasts
about $30^{\circ}$ rotations when it is turned on or off. 
Our robot starts from both the left (off state) and the right side (on state) of the knob.
The \textbf{speaker}
has a tiny cylindrical knob that decreases in its diameter from $13.1 mm$
to $9.1 mm$ with height of $12.8 mm$ and requires $30^{\circ}$ degree rotation.
%However, given that PR2 fingertips with silicon covers measure $23 mm$ and are parallel plates,
%grasping a tiny $9.1 mm$ knob %of speaker
However, since PR2 fingertips are parallel plates and measure $23 mm$ with silicon covers,
grasping a $9.1 mm$ knob %of speaker
results in drastically different sensor readings at every execution of the task.
The \textbf{desk fan}
has a rectangular knob with a width of $25.1 mm$ and a large surface area.
It has a two-step detent control with a click that lasts $45^{\circ}$ degree rotation and 
has a narrow stoppable window of about $\pm 20^{\circ}$ degrees.

The stirrer and the speaker can both be rotated clockwise and counterclockwise
and have a wall at both ends.
The desk fan has three stoppable points (near $0^{\circ}$, $45^{\circ}$, and $90^{\circ}$) 
to adjust fan speed 
and can get stuck in-between if a rotation is not enough or exceeds a stopping point.

%%%%%%%%%%%%% COLLECTION %%%%%%%%%%%%%%%%
Each object is provided with 
a nominal plan with multiple phases, 
each defined as a sequence of smoothly interpolated waypoints consisting 
of end-effector position and orientation along with gripper actions of grasping
similar to \cite{sung_robobarista_2015}.
For each of the objects, we collected at least $25$ successes and $25$ failures.
The success cases only includes rotations that resulted in successful transition
of states of objects (e.g. from off to on state).
The failures include slips, excessive rotations beyond acceptable range, 
rotation even after hitting a wall, and near breaking of the knob.
There also exists less dramatic failures such as insufficient rotations.
Especially for the desk fan, 
if a rotation results in two clicks beyond the first stopping point,
it is considered a failure.
Each data sequence consists of a sequence of trajectory (phases) 
as well as tactile sensor signal after an execution of each waypoint.

To label the reward for each sequence,
an external camera with a microphone was placed nearby the object.
By reviewing the audio and visually inspecting haptic signal afterwards,
an expert labeled the timeframe that each sequence succeeded or failed.
These extra recordings were only used for labeling the rewards,
and such input is not made available to the robot during our experiments.
For sequences that turned the knob past the successful
stage but did not stop the rotation, only negative rewards were given.

Among multiple phases of a nominal plan,
which includes pre-grasping and post-interaction trajectories,
we focus on three phases (before-rotation/rotation/stopped).
These phases occur after grasping and success is determined 
by ability to correctly rotate and detect when to shift to the final phase.

%% file: 4_results_baseline.tex
\subsection{Baselines}
\label{sec:baselines}
\vspace*{\subsectionReduceBot}

We compare our model against several baseline methods on our dataset and in our robotic experiment.
Since most of the related works are applied to problems in different domains,
we take key ideas (or key structural differences) from relevant works
and fit them to our problem.

\noindent
1) \emph{Chance:}
    It follows a nominal plan and makes a transition between phases by randomly selecting the amount
    of degree to rotate a knob without incorporating haptic feedback.
    
\noindent
2) \emph{Non-recurrent Recognition Network:}
     Similar to \cite{watter2015embed}, we take non-recurrent deep neural network
     of only observations without actions.
     However, it has access to a short history in a sliding window of haptic signal 
     at every frame.
     For control, we apply the same Q-learning method as our full model.

\noindent
3) \emph{Recurrent Network as Representation:}
    Similar to \cite{lenz2015deepmpc},
    we directly train a recurrent network to predict future haptic signals.
    %rather than training to learn a latent representation as we did in our model.
    At each time step $t$, a LSTM network takes concatenated observation $o_t$ and previous action $a_t$
    as input, and 
    the output of the LSTM is concatenated with $a_{t+1}$ to predict $o_{t+1}$.
    However, while \cite{lenz2015deepmpc} relies on hand-coded MPC cost function to choose an action,
    we apply same Q-learning that was applied to our full model.
    For haptic prediction experiment, 
    transitions happen by taking the output of the next time step
    as input to the next observation.

\noindent
4) \emph{Our Model without Exploration:}
    During the final deep Q-Learning (Sec.~\ref{sec:dql}) stage, 
    it skips the exploration step that uses a learned transition model
    and only uses sequences of representation from the recognition network.

%% file: 5_conclusion.tex
%\vspace*{-0.1in}
\section{Conclusion} 
\label{sec:conclusion}
%\vspace*{-0.1in}
\vspace*{\sectionReduceBot}

In this work, we present a novel framework for learning to represent
haptic feedback of an object that requires sense of touch.
We model such tasks as partially observable model
with its generative model parametrized by neural networks.
To overcome intractability of computing posterior,
variational Bayes method allows us to approximate posterior
with a deep recurrent recognition network consisting of two LSTM layers.
Using a learned representation of haptic feedback,
we also introduce a Q-learning method that is able to learn optimal control
without access to simulator in learned latent state space utilizing
only prior experiences and learned generative model for transition.
We evaluate our model on a task of rotating a knob 
until it clicks 
against several baseline.
With more than 200 robotic experiments on the PR2 robot, 
we show that our model is able to successfully manipulate knobs that click
while predicting future haptic signals.

%% file: 6_appendix.tex
%\vspace*{-.1in}
\section*{APPENDIX}
%\vspace*{-.05in}
\vspace*{\sectionReduceBot}

%\vspace*{-.05in}
\subsection{Lowerbound Derivation}
\label{sec:derivation}
\vspace*{\subsectionReduceBot}

%To continue our derivation of the lower bound on conditional 
%log-likelihood from Sec.~\ref{sec:variational}.
%The second term of equation~\ref{eq:recon}:
To continue our derivation of the lower bound from Sec.~\ref{sec:variational}.
The second term of equation~\ref{eq:recon}:
\begin{align*}
\mathbb{E}_{q_\phi(\vs|\vo,\vr,\va)} & \big[ log \; p_\theta(\vo,\vr|\vs,\va) \big] \\
&= \mathbb{E}_{q_\phi(\vs|\vo,\vr,\va)} \big[ log \; p_\theta(\vo|\vs) + log \; p_\theta(\vr|\vs) \big] \\
&\approx \frac{1}{L} \sum_{l=1}^{L} \big[ log \; p_\theta(\vo|\vs^{(l)}) + log \; p_\theta(\vr|\vs^{(l)}) \big] \\
&= \frac{1}{L} \sum_{l=1}^{L} \sum_{t=1}^{T}  \big[ log \; p_\theta(o_t|s_t^{(l)}) + log \; p_\theta(r_t|s_t^{(l)}) \big] \\
& \;\;\;\; \text{ where } \vs^{(l)}=q_{\phi,\mu} + \epsilon^{(l)} q_{\phi,\Sigma} \text{ and } \epsilon^{(l)} \sim p(\epsilon)
%& \;\;\;\; \text{ where } \vs^{(l)}=g_\phi(\epsilon^{(l)}, \vo, \vr, \va) \text{ and } \epsilon^{(l)} \sim p(\epsilon)
\end{align*}
Reparametrization trick (\cite{kingma2013auto,rezende2014stochastic}) at last step 
samples from the inferred distribution by a recognition network $q_\phi$.

And, for the first term from equation~\ref{eq:recon}:
\begin{align*}
&D_{KL}\big( q_\phi(\vs|\vo,\vr,\va) || p_\theta(\vs|\va) \big) \\
&= \int_{s_1} \dotsb \int_{s_T} q_\phi(\vs|\vo,\vr,\va) \bigg[ log \frac{q_\phi(\vs|\vo,\vr,\va)}{p_\theta(\vs|\va)} \bigg] \\
&= D_{KL}\big( q_\phi(s_1|\vo,\vr,\va) || p(s_1) \big) \\
& \quad + \sum_{t=2}^{T} \mathbb{E}_{s_{t-1} \sim q_\phi(s_{t-1}|\vo,\vr,\va)} \big[ \\ 
& \qquad\qquad\qquad D_{KL}\big(q_\phi(s_t|s_{t-1},\vo,\vr,\va)||p(s_t|s_{t-1},a_{t-1})\big) \big] \\
& \text{using reparameterazation trick again,} \\
&= D_{KL}\big( q_\phi(s_1|\vo,\vr,\va) || p(s_1) \big) \\
& \quad + \sum_{t=2}^{T} \frac{1}{L}\sum_{l=1}^{L} \big[ 
D_{KL}\big(q_\phi(s_t|s_{t-1},\vo,\vr,\va)||p(s_t|s_{t-1}^{(l)},a_{t-1})\big) \big] \\
& \;\;\;\; \text{ where } s_{t-1}^{(l)}=q_{\phi,t-1,\mu} + \epsilon^{(l)} q_{\phi,t-1,\Sigma} \text{ and } \epsilon^{(l)} \sim p(\epsilon) 
%& \;\;\;\; \text{ where } s_{t-1}^{(l)}=g_\phi(\epsilon^{(l)}, \vo, \vr, \va) \text{ and } \epsilon^{(l)} \sim p(\epsilon) 
\end{align*}
Combining these two terms, we arrive at
equation~\ref{eq:recon_final}.

We do not explain each step of the derivation at length since 
similar ideas behind the derivation can be found at \cite{krishnan2015deep}
although exact definition and formulation are different.